\documentclass{article}

\usepackage[final]{bdl_2018}

\usepackage[utf8]{inputenc} 
\usepackage[T1]{fontenc}    
\usepackage{hyperref}       
\usepackage{url}            
\usepackage{booktabs}       
\usepackage{amsfonts}       
\usepackage{nicefrac}       
\usepackage{microtype}      
\usepackage{graphicx}
\usepackage{float}
\usepackage{subcaption}
\usepackage{amsmath}
\usepackage{amssymb}
\DeclareMathOperator{\E}{\mathbb{E}}

\title{BAR: Bayesian Activity Recognition using variational inference}

\author{
Ranganath Krishnan\\ranganath.krishnan@intel.com\And Mahesh Subedar\\mahesh.subedar@intel.com\And Omesh Tickoo\\omesh.tickoo@intel.com \\
\and\\
Intel Labs\\
Hillsboro, OR (USA) \\
}

\begin{document}
	\bibliographystyle{unsrt}

\maketitle

\begin{abstract}
Uncertainty estimation in deep neural networks is essential for designing reliable and robust AI systems. Applications such as video surveillance for identifying suspicious activities are designed with deep neural networks (DNNs), but DNNs do not provide uncertainty estimates. Capturing reliable uncertainty estimates in safety and security critical applications will help to establish trust in the AI system. Our contribution is to apply Bayesian deep learning framework to visual activity recognition application and quantify model uncertainty along with principled confidence. We utilize the stochastic variational inference technique while training the Bayesian DNNs to infer the approximate posterior distribution around model parameters and perform Monte Carlo sampling on the posterior of model parameters to obtain the predictive distribution.
We show that the Bayesian inference applied to DNNs provide reliable confidence measures for visual activity recognition task as compared to conventional DNNs.
We also show that our method improves the visual activity recognition precision-recall AUC by 6.2\% compared to non-Bayesian baseline. We evaluate our models on Moments-In-Time (MiT) activity recognition dataset by selecting a subset of in- and out-of-distribution video samples.
\end{abstract}

\section{Introduction}
Activity recognition is an active area of research with multiple approaches depending on the application domain and the types of sensors~\citep{bulling2014tutorial, hershey2017cnn, caba2015activitynet}.  In recent years, the DNN models applied to the visual activity recognition task outperform the earlier methods based on Gaussian Mixture Models and Hidden Markov Models~\citep{piyathilaka2013gaussian, kim2010human} using handcrafted features. The availability of large datasets~\citep{kay2017kinetics,monfort2018moments,abu2016youtube} for visual analysis tasks has enabled the use of DNNs for activity recognition task. Vision-based activity recognition methods typically apply a combination of spatiotemporal features to classify the activities. Single frame based activity recognition methods usually apply DNN models trained on ImageNet dataset~\citep{szegedy2017inception} to extract the spatial features. The temporal dynamics for activity recognition is typically modeled either by using a separate temporal sequence modeling such as variants of RNNs~\citep{wang2016temporal,zhou2017temporal} or by applying 3D ConvNets~\citep{tran2015learning}, which extend 2D ConvNets to the temporal dimension. In this work, we focus on the visual activity recognition applied to the trimmed video samples using the 3D ConvNet (C3D) architecture~\citep{hara3dcnns}

Probabilistic Bayesian models provide principled ways to gain insight about data and capture reliable uncertainty estimates in predictions. Bayesian deep learning has allowed bridging DNNs and probabilistic Bayesian theory to leverage the strengths of both methodologies~\citep{neal2012bayesian,gal2016uncertainty}. 
Conventional DNNs are trained to obtain the maximum likelihood estimates and they tend to disregard uncertainty around the model parameters that eventually leads to predictive uncertainty.
DNNs may fail in the case of noisy or out-of-distribution data, leading to overconfident decisions that could be erroneous as SoftMax probability does not capture overall model confidence. Instead, it represents relative probability that an input is from a particular class compared to the other classes.

Bayesian deep learning framework with Monte Carlo (MC) dropout technique has been used in visual scene understanding applications including camera relocalization \citep{kendall2016modelling}, semantic segmentation \citep{kendall2015bayesian}, and depth regression \citep{kendall2017multi}. In this work, we propose a Bayesian confidence measure applied to visual activity recognition task using the variational inference technique. To the best of our knowledge, this is the first research effort that applies Bayesian deep learning framework with variational inference for activity recognition task to capture reliable uncertainty measures.

\section{Background}
\label{gen_inst}
Bayesian neural networks (BNNs) offer a probabilistic interpretation of deep learning models by placing distributions over the model parameters. Bayesian inference can be applied to estimate the predictive distribution by propagating over the model likelihood and marginalizing over the learned posterior parameter distribution. BNNs also help in regularization by introducing distribution over network weights, capturing the posterior uncertainty around the neural network parameters. This allows transferring inherent DNN uncertainty from the parameter space to the predictive uncertainty of the unseen data.

Given training dataset $D=\{x,y\}$ with inputs $x = {x_1, . . . , x_N}$ and their corresponding outputs $y = {y_1, . . . , y_N}$, in parametric Bayesian settings we would like to infer a distribution over weights $w$ as a function $y = f_w(x)$ that represents structure of the DNN model.
With the posterior for model parameters inferred during Bayesian neural network training, we can predict the output for a new data point by propagating over the model likelihood $p(y|x,w)$ while drawing samples from the learned parameter posterior $p(w|D$). Equation~\ref{eq:pred_dist} below shows predictive distribution of output $y^*$ given new input $x^*$:
\begin{equation}
p(y^*\,|\,x^*,D) = \int p(y^*\,|\,x^*,w)\,\,p(w\,|\,D) dw
\label{eq:pred_dist}
\end{equation}

In Bayesian neural networks, some of the techniques to achieve tractable inference include: (i) Markov Chain Monte Carlo (MCMC) sampling based probabilistic inference \citep{neal2012bayesian,welling2011bayesian} (ii) Variational inference techniques to infer the tractable approximate posterior distribution around model parameters \citep{graves2011practical,ranganath2013black,blundell2015weight} and (iii) Monte Carlo (MC) dropout approximate inference \citep{gal2016dropout}. In our work, we use variational inference approach to infer the approximate posterior distribution around the model parameters.

Variational inference~\citep{blei2017variational} is an active area of research in Bayesian deep learning, which uses gradient based optimization. This technique approximates complex probability distribution $p(w|D)$ with a simpler distribution $q_\theta (w)$, parameterized by variational parameters $\theta$ while minimizing the Kullback-Leibler (KL) divergence.
Minimizing the KL divergence is equivalent to maximizing the log evidence lower bound \citep{bishop2006pattern,gal2016dropout}.
\begin{equation}
\mathcal{L_{VI}} := \int q_\theta(w)\,log\,p(y\,|\,x,w)\,dw - KL[q_\theta(w)\,||\,p(w)]
\end{equation}


We evaluate the model uncertainty using Bayesian active learning by disagreement (BALD) \citep{houlsby2011bayesian} for the visual activity recognition task. BALD quantifies mutual information between parameter posterior distribution and predictive distribution, which captures model uncertainty, as shown in Equation~\ref{eq:mutual information}.

\begin{equation}
BALD := H(y^*\,|\,x^*, D)-\E_{q_\theta (w)}[H(y^*\,|\,x^*, w)]\\
\label{eq:mutual information}
\end{equation}
where, $H(y^*\,|\,x^*, D)$ is the predictive entropy given by:
\begin{equation}
H(y^*\,|\,x^*, D)=-\sum_{i=0}^{K-1}p_{i\mu}\,*\,log\,p_{i\mu}\\
\label{eq:pred_entropy}
\end{equation}
and $p_{i\mu}$ is predictive mean probability of $i^{th}$ class from $T$ Monte Carlo samples.

\section{Bayesian DNN architecture}
\label{archiecture}
\begin{figure}
\centering
\includegraphics[width=0.9\textwidth]{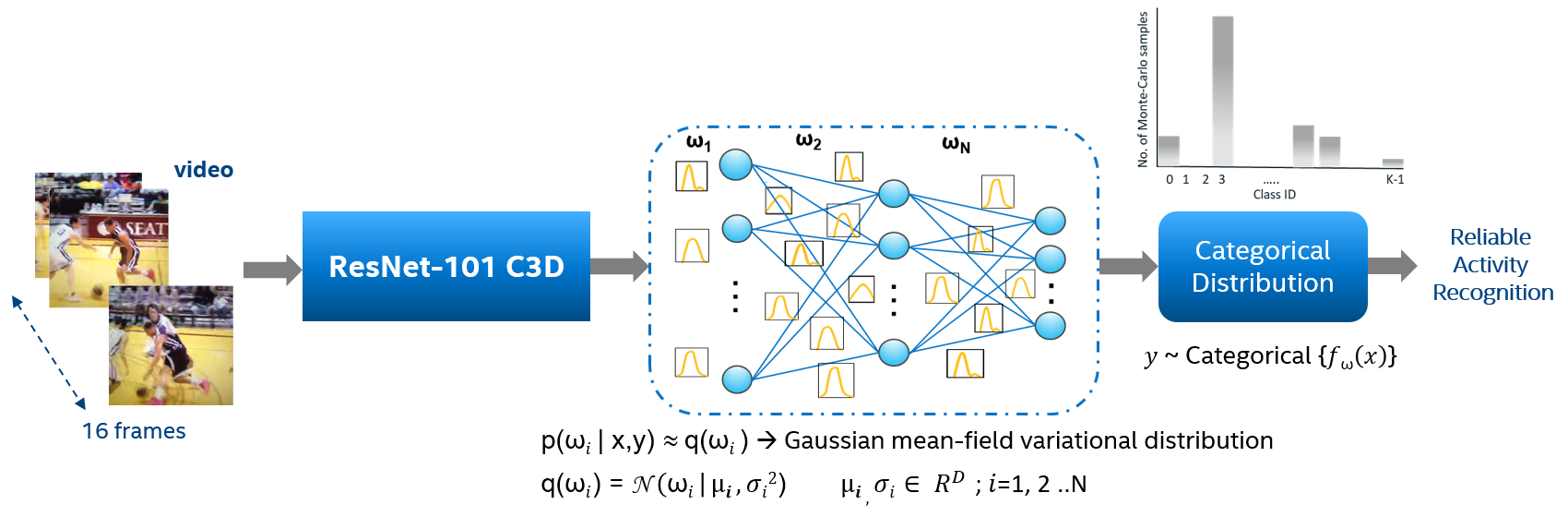}
\caption{Vision-based activity recognition using ResNet101 C3D architecture. Bayesian variational inference is applied to the final three fully connected layers. Monte Carlo sampling on the posterior of model parameters provides the predictive distribution.}
\label{fig:blk_diagram}
\end{figure}
We use ResNet-101 C3D~\citep{hara3dcnns} architecture and replace the final layer with three fully connected variational layers followed by categorical distribution, as shown in Figure~\ref{fig:blk_diagram}. The weights and bias parameters in the fully connected variational layers are modeled through mean-field normal distribution, and the network is trained using Bayesian variational inference based on KL divergence~\citep{ranganath2013black,blundell2015weight}. We use Flipout~\citep{wen2018flipout}, which is an efficient method that decorrelates the gradients within a mini-batch by implicitly sampling pseudo-independent weight perturbations for each input. We perform 40 stochastic forward passes on the final three fully connected variational layers with Monte Carlo sampling on the weight and bias posterior distributions, while the remaining layers of the network are considered to be deterministic. 
We compare the proposed architecture with MC dropout approximate Bayesian inference~\citep{gal2016dropout} method. We analyze the confidence measure and model uncertainty estimates for true (correct) and false (incorrect) predictions obtained from Bayesian DNN models.

For the comparison with the non-Bayesian baseline, we maintain the same model depth as the Bayesian DNN model and use three deterministic fully connected final layers for the non-Bayesian DNN model. The dropout layer is used after every fully connected layer to avoid over-fitting of the model. In the rest of the document, we refer the non-Bayesian DNN model as simply the DNN model.


\begin{figure} [b]
\centering
\includegraphics[width=0.42\linewidth]{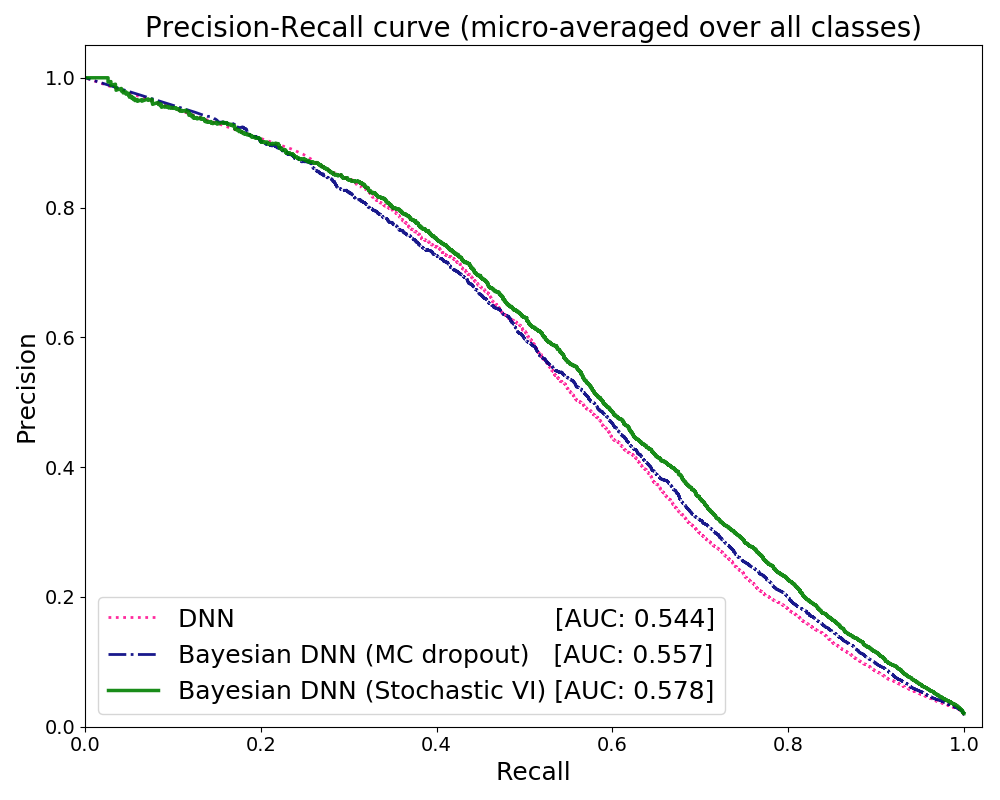}
\includegraphics[width=0.42\linewidth]{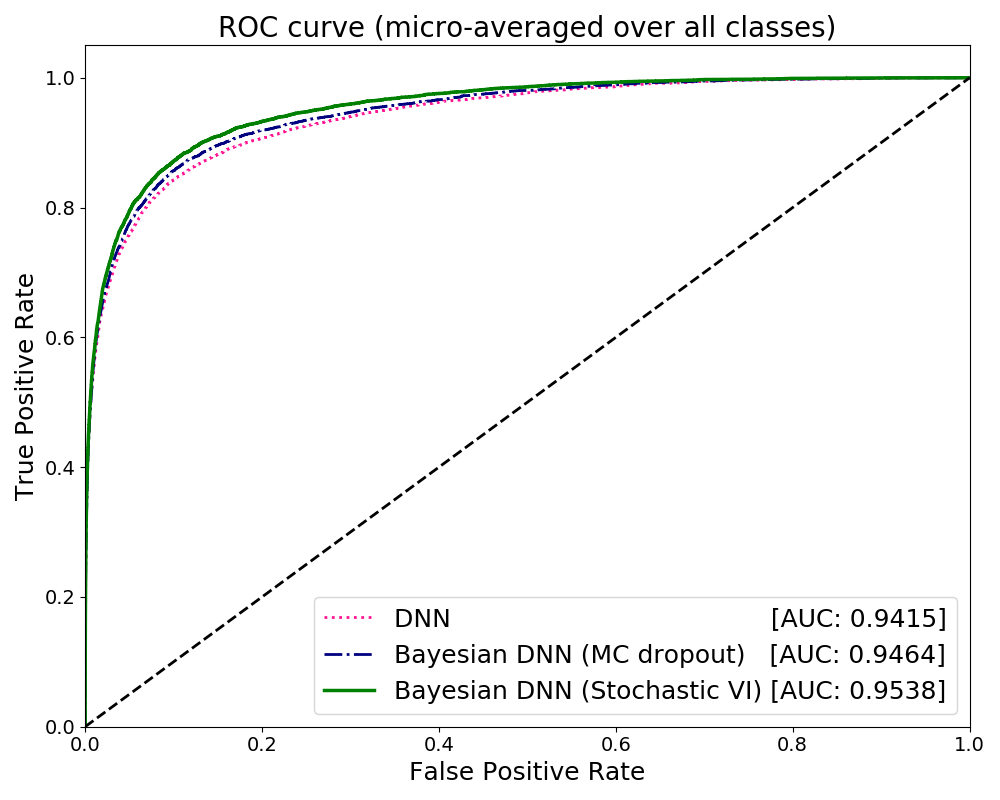}
\caption{Comparison of DNN and Bayesian DNN visual activity recognition results using Precision-Recall and ROC plots.}
\label{fig:PrecxRecall Curve}
\end{figure}

\section{Results}
We evaluate our models on the MiT video activity recognition dataset \citep{monfort2018moments} and train our model on a subset of 54 classes, which are used as in-distribution data. We use a different subset of 54 classes as out-of-distribution data. The subset of 54 classes for each category are selected after subjective evaluation to confirm the selected classes are not ambiguous and they fall into two distinct distribution of classes.

\begin{table}[t]
\begin{center}
\caption{Comparison of accuracies for DNN, Bayesian DNN MC Dropout and Stochastic Variational Inference (Stochastic~VI) models applied to the subset of MiT dataset (in-distribution classes).}
\begin{tabular}{|l|c|c|}
\hline
\textbf{Model} & \textbf{Top1 (\%)}  & \textbf{Top5 (\%)} \\\hline
DNN Model & 53.90 & 81.25 \\\hline
Bayesian DNN (MC Dropout) Model & 53.40 & 80.81 \\\hline
Bayesian DNN (Stochastic VI) Model &\textbf{54.10} & \textbf{81.40} \\
\hline
\end{tabular}
\label{tab:Accuracy}
\end{center}
\end{table}

\begin{figure}[t]
\centering
\begin{subfigure}[t]{0.33\linewidth}
\centering
\includegraphics[width=0.99\textwidth]{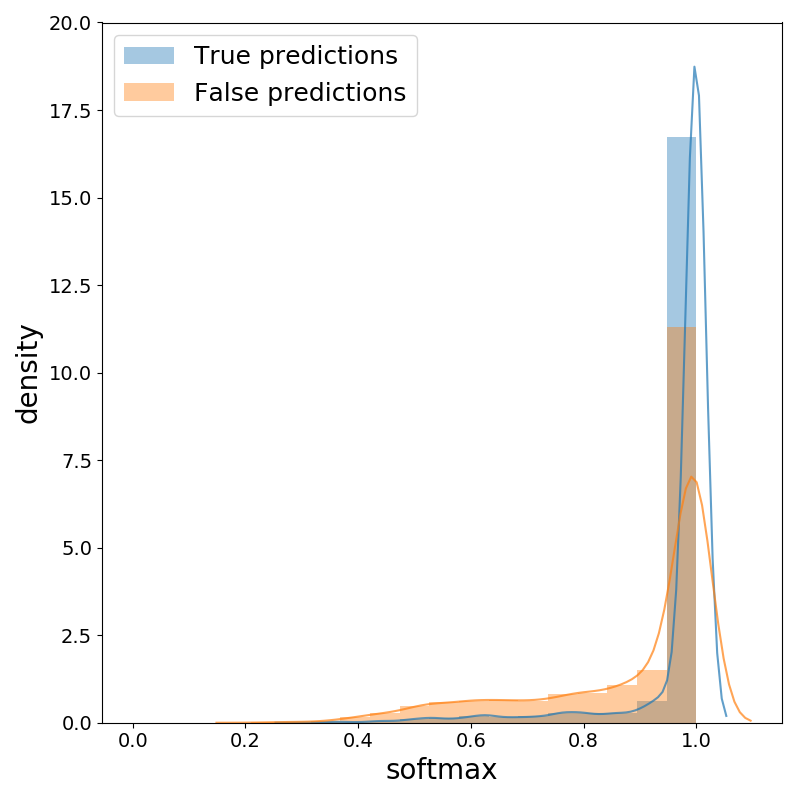}
\caption{\small{Baseline DNN}}
\end{subfigure}%
\begin{subfigure}[t]{0.33\textwidth}
\centering
\includegraphics[width=0.99\textwidth]{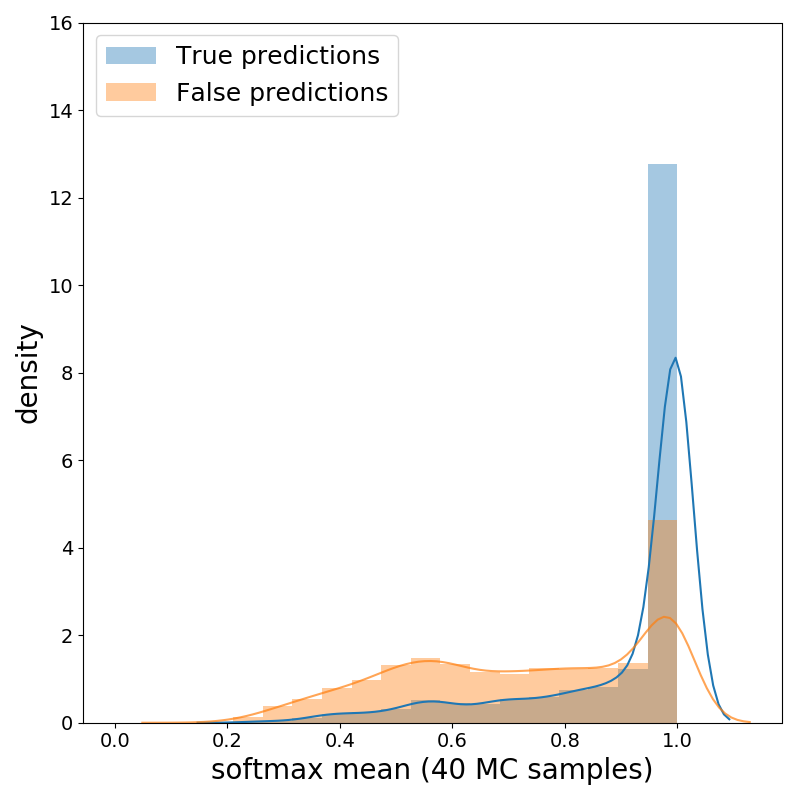}
\caption{\small{Bayesian DNN (MC Dropout)}}
\end{subfigure}
\begin{subfigure}[t]{0.33\textwidth}
\centering
\includegraphics[width=0.99\textwidth]{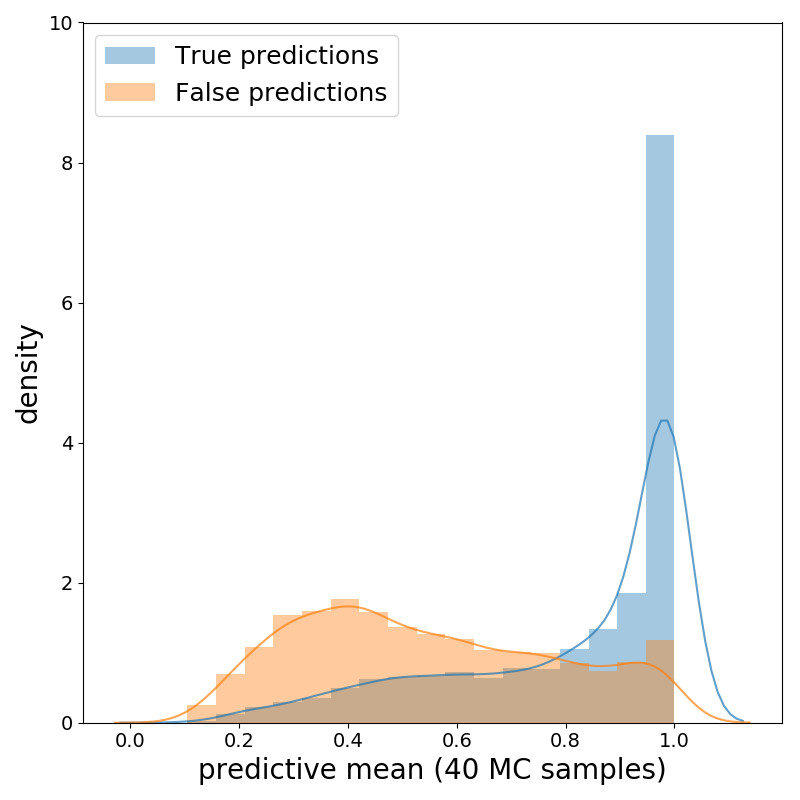}
\caption{\small{Bayesian DNN (Stochastic-VI)}}
\end{subfigure}
\caption{\small{Density histogram of confidence measures for true (correct) and false (incorrect) predictions using DNN and Bayesian DNN models: A distribution skewed towards right (near 1.0 on x-axis) indicates the model has higher confidence in predictions than the distribution skewed towards left. [The density histogram is a histogram with area normalized to one. Plots are overlaid with kernel density curves for better readability.]}}
\label{fig:ConfMeasuresTrueFalse}
\end{figure}
\begin{figure}[b]
\centering
\begin{subfigure}[t]{0.33\linewidth}
\centering
\includegraphics[width=0.99\textwidth]{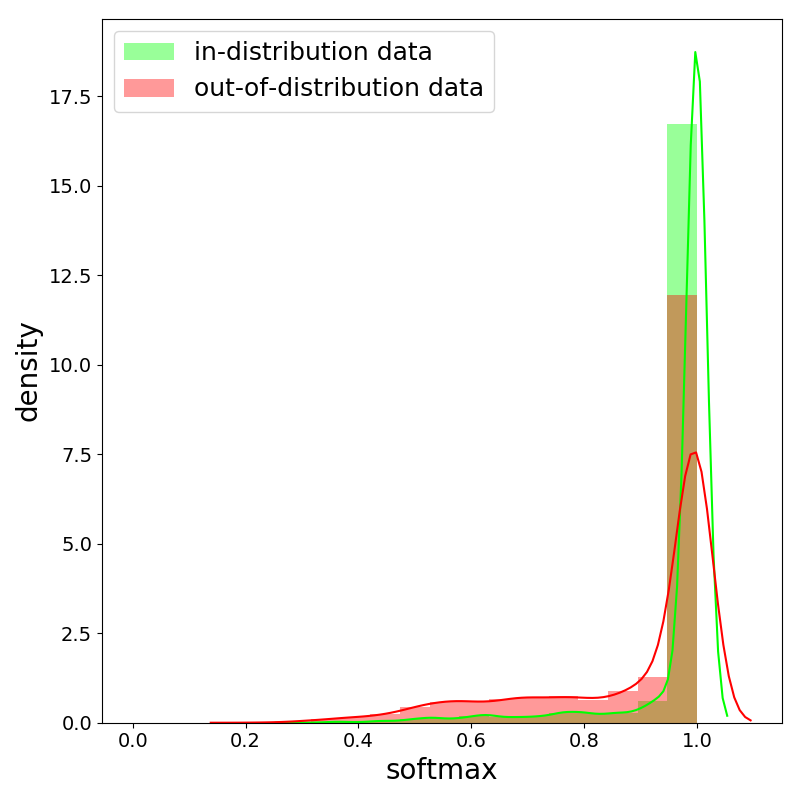}
\caption{\small{Baseline DNN}}
\end{subfigure}%
\begin{subfigure}[t]{0.33\textwidth}
\centering
\includegraphics[width=0.99\textwidth]{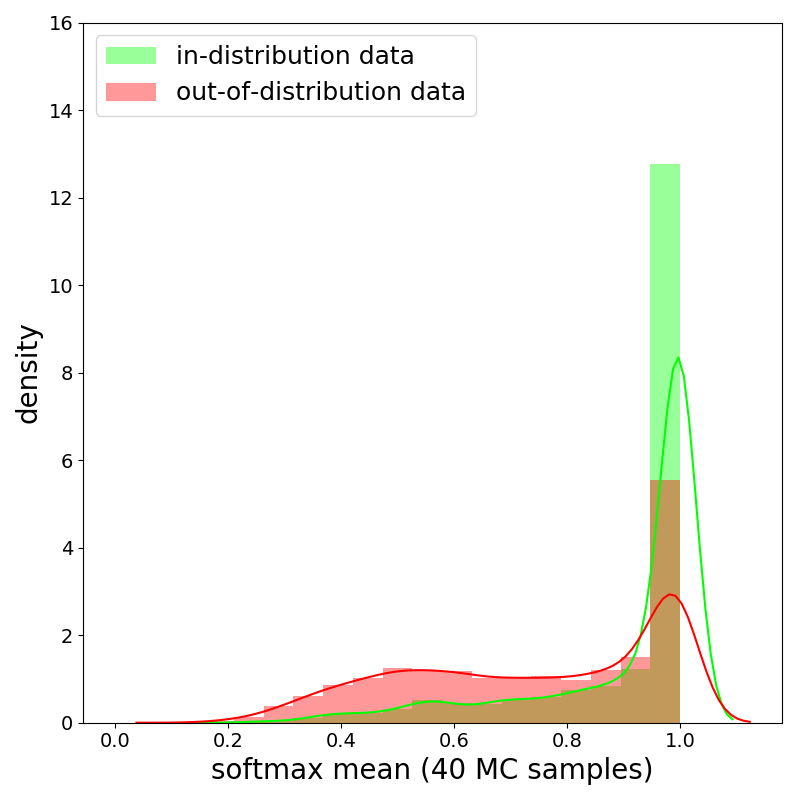}
\caption{\small{Bayesian DNN (MC Dropout)}}
\end{subfigure}
\begin{subfigure}[t]{0.33\textwidth}
\centering
\includegraphics[width=0.99\textwidth]{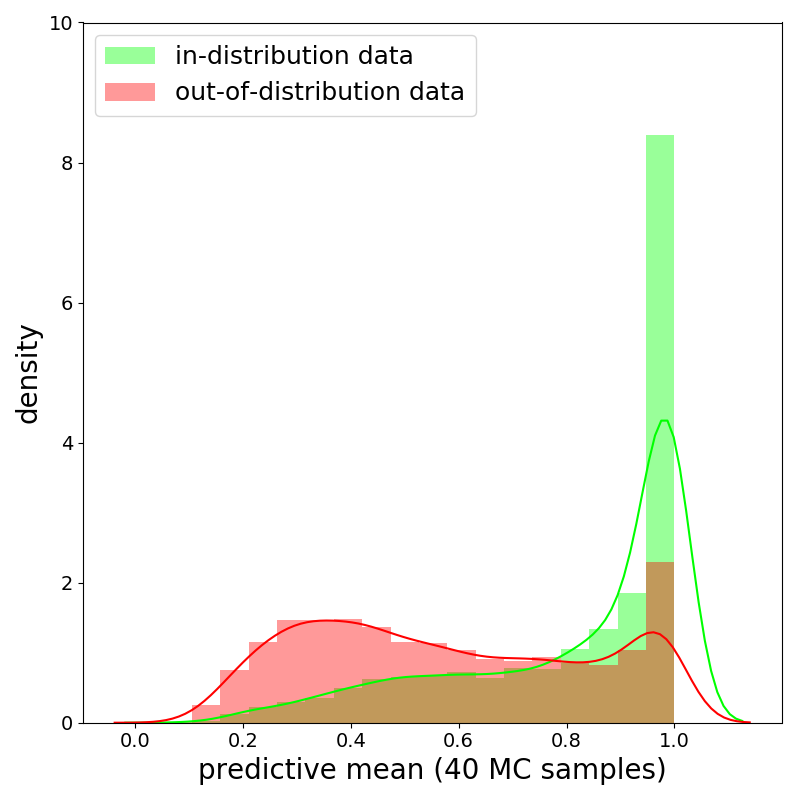}
\caption{\small{Bayesian DNN (Stochastic-VI)}}
\end{subfigure}
\caption{\small{Density histogram of confidence measures for in- and out-of-distribution samples using DNN and Bayesian DNN models. [The density histogram is a histogram with area normalized to one. Plots are overlaid with kernel density curves for better readability.]}}
\label{fig:ConfMeasuresInOut}
\end{figure}

\newcommand{\mysize}{0.35}
\newcommand{\myhspace}{0.1}
\newcommand{\imagewidth}{0.95}
\begin{figure}[t]
\centering
\begin{subfigure}[t]{\mysize\linewidth}
\centering
\includegraphics[width=\imagewidth\textwidth]{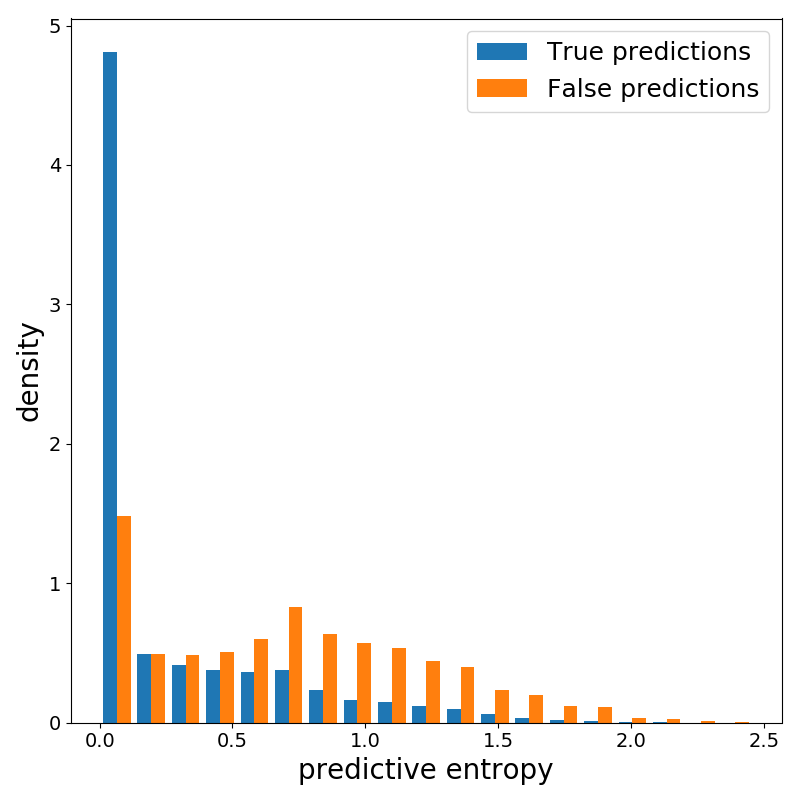}
\caption{\small{Predictive Entropy (MC Dropout)}}
\end{subfigure}%
\hspace{\myhspace\textwidth}
\begin{subfigure}[t]{\mysize\textwidth}
\centering
\includegraphics[width=\imagewidth\textwidth]{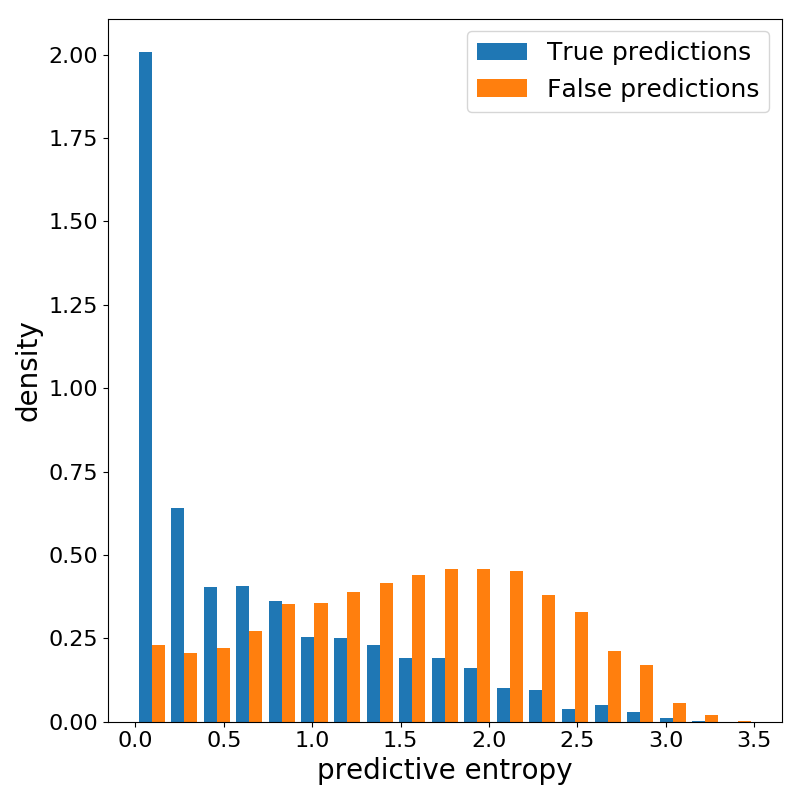}
\caption{\small{Predictive Entropy (Stochastic VI)}}
\end{subfigure}
\par\bigskip
\begin{subfigure}[t]{\mysize\textwidth}
\centering
\includegraphics[width=\imagewidth\textwidth]{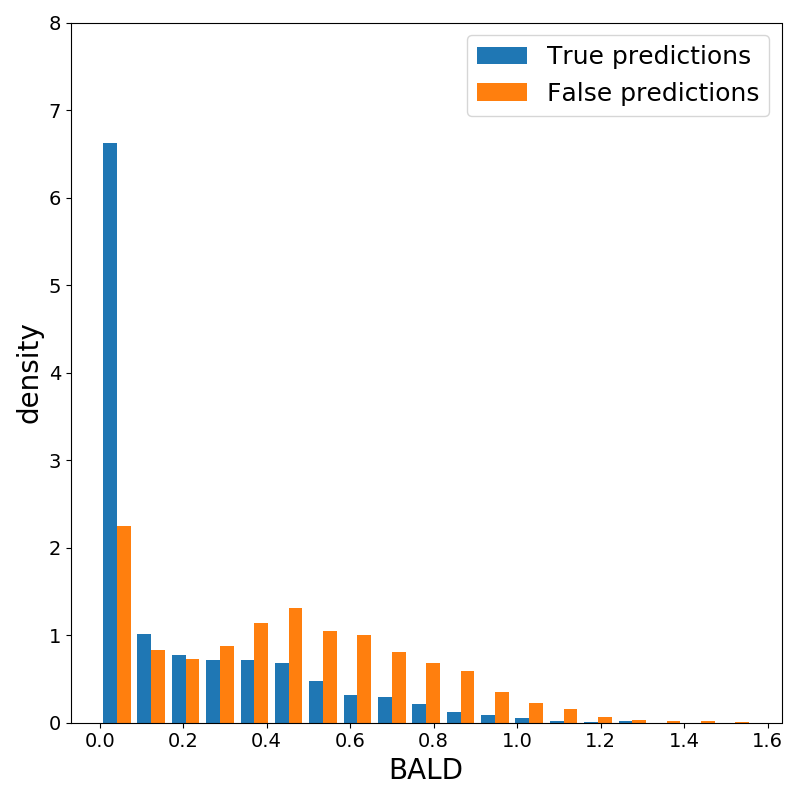}
\caption{\small{BALD (MC Dropout)}}
\end{subfigure}
\hspace{\myhspace\textwidth}
\begin{subfigure}[t]{\mysize\textwidth}
\centering
\includegraphics[width=\imagewidth\textwidth]{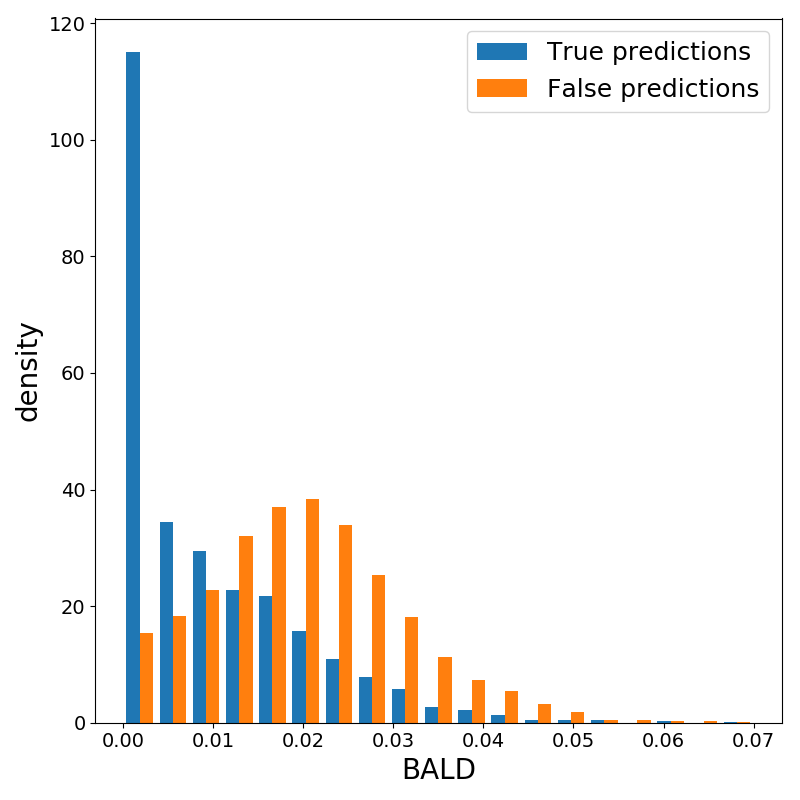}
\caption{\small{BALD (Stochastic VI)}}
\end{subfigure}
\caption{\small{Density histogram of uncertainty measures from Bayesian DNN  MC dropout and stochastic variational inference (Stochastic VI) models for true (correct) and false (incorrect) predictions from in-distribution samples. [The density histogram is a histogram with area normalized to one.]}}
\label{fig:uncertnityest}
\end{figure}

In Figure~\ref{fig:PrecxRecall Curve}, comparison of precision-recall (left) and ROC (right) plots are presented for the three models. The Bayesian DNN stochastic variational inference (Stochastic VI) results achieve an improvement of 6.2\% in precision-recall AUC over the non-Bayesian DNN model and an improvement of 3.63\% over the MC dropout approximate Bayesian inference model. The ROC AUC curve for Bayesian DNN stochastic variational inference shows an improvement of 1.3\% over the DNN results. The improvement in AUC for the stochastic variational inference model applied to activity recognition task is attributed to better measure of the confidence estimates obtained from the predictive distribution through Monte Carlo sampling. The classification accuracy for MiT in-distribution validation data is presented in Table~\ref{tab:Accuracy}. The accuracies for proposed stochastic variational inference model is similar to the DNN model.

In Figure~\ref{fig:ConfMeasuresTrueFalse}, the density histograms of confidence measure for true (correct) and false (incorrect) predictions is presented. The density histogram is a histogram with area normalized to one.  The confidence measure for the conventional DNN is the SoftMax probabilities used for the predictions. The mean of categorical predictive distribution obtained from Monte Carlo sampling provides the confidence measure for Bayesian DNNs. The DNN model density histogram shows a peak near higher confidence values for the true predictions, similar to the Bayesian DNN models. The DNN SoftMax probability measure (Figure~\ref{fig:ConfMeasuresTrueFalse}~(a)) is skewed towards higher probability values for the false predictions. On the contrary, predictive mean probabilities obtained from the Bayesian DNN Monte Carlo dropout (Figure\ref{fig:ConfMeasuresTrueFalse}~(b)) and stochastic variational inference (Figure~\ref{fig:ConfMeasuresTrueFalse}~(c)) models provide more reliable confidence measure indicating lower confidence values for the false predictions.

In Figure~\ref{fig:ConfMeasuresInOut}, density histogram of confidence measures for in- and out-of-distribution samples is presented. All the models show higher confidence values for in-distribution data. For out-of-distribution data, DNN SoftMax probability values (Figure~\ref{fig:ConfMeasuresInOut}~(a)) show overconfident measure for the erroneous predictions, whereas Bayesian DNNs (Figure~\ref{fig:ConfMeasuresInOut} (b) \& (c)) models show lower confidence for out-of-distribution data. Stochastic variational inference approach gives better results than MC dropout approach showing a more pronounced peak towards lower value for out-of-distribution samples.

Bayesian DNN models quantify uncertainty which is beneficial to identify false predictions and out-of-distribution samples. We compare predictive entropy and BALD~\citep{houlsby2011bayesian} uncertainty measures for Bayesian Monte Carlo dropout and stochastic variational inference models. In Figure~\ref{fig:uncertnityest}, the density histogram of uncertainty estimates 
for true and false predictions using the MiT dataset is presented. The uncertainty estimates for both the models are skewed towards higher values for the false predictions, but are skewed towards lower uncertainty for the true predictions. This indicates the models reliably capture the uncertainty associated with the predictions.

We also compare the uncertainty estimates using UCF101~\citep{soomro2012ucf101} visual action recognition dataset consisting of 101 activity classes. The comparison of uncertainty measures for UCF101 dataset as in-distribution samples and the MiT dataset as out-of-distribution samples is shown in Figure~\ref{fig:uncertnityestUCF101}. Predictive entropy and BALD uncertainty measures demonstrate a clear separation of in- and out-of-distribution samples. The stochastic variational inference approach shows better separation of the peaks for in- and out-of-distribution samples (Figure~\ref{fig:uncertnityestUCF101}~(b)~\&~(d)) compared to the Monte Carlo dropout method (Figure~\ref{fig:uncertnityestUCF101}~(a)~\&~(c)).

These results indicate the Bayesian DNNs for activity recognition task provide better confidence measure compared to the non-Bayesian DNN model, along with uncertainty estimates to reliably identify out-of-distribution data.




\renewcommand{\mysize}{0.35}
\renewcommand{\myhspace}{0.1}
\renewcommand{\imagewidth}{0.95}
\begin{figure}
\centering
\begin{subfigure}[t]{\mysize\linewidth}
\centering
\includegraphics[width=\imagewidth\textwidth]{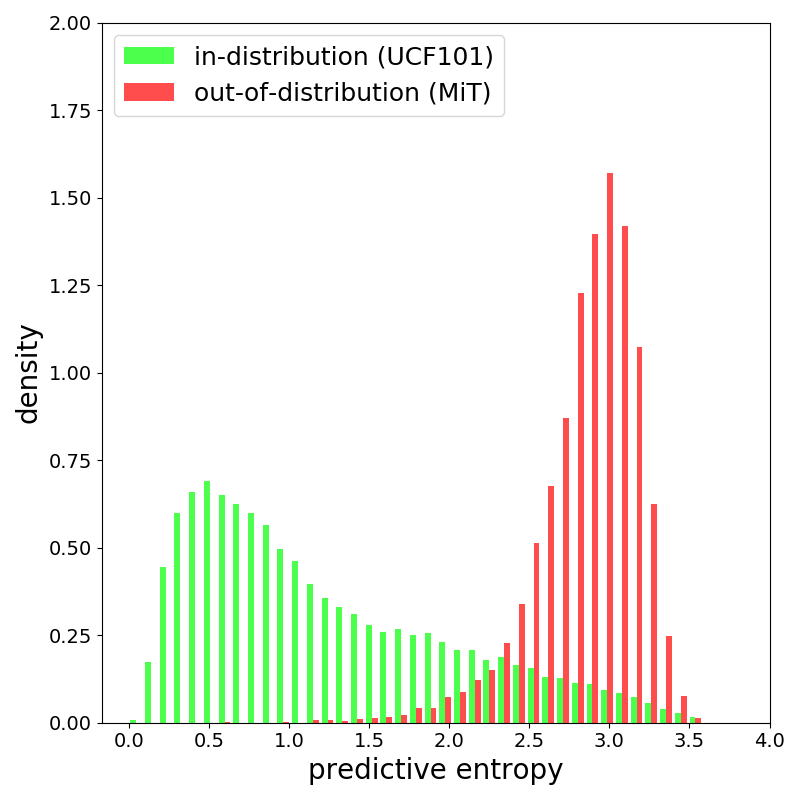}
\caption{\small{Predictive Entropy (MC Dropout)}}
\end{subfigure}%
\hspace{\myhspace\textwidth}
\begin{subfigure}[t]{\mysize\textwidth}
\centering
\includegraphics[width=\imagewidth\textwidth]{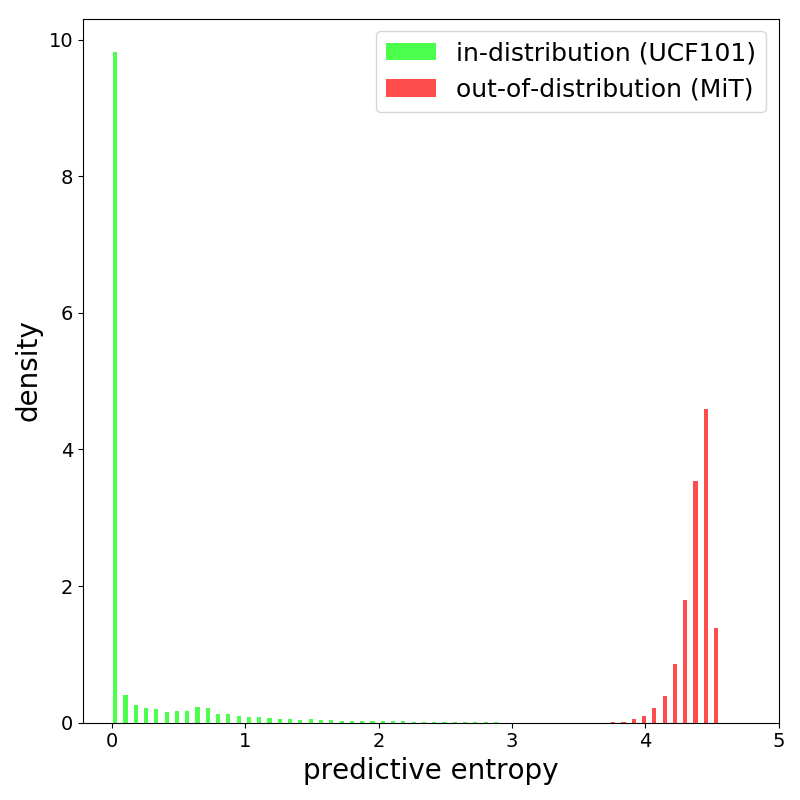}
\caption{\small{Predictive Entropy (Stochastic VI)}}
\end{subfigure}
\par\bigskip
\begin{subfigure}[t]{\mysize\textwidth}
\centering
\includegraphics[width=\imagewidth\textwidth]{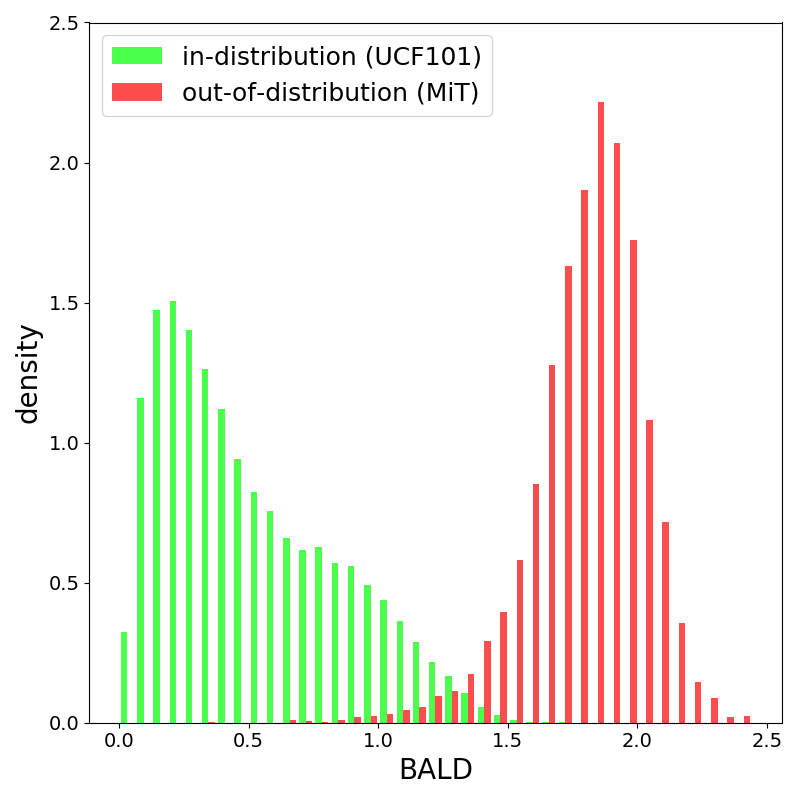}
\caption{\small{BALD (MC Dropout)}}
\end{subfigure}
\hspace{0.1\textwidth}
\begin{subfigure}[t]{\mysize\textwidth}
\centering
\includegraphics[width=\imagewidth\textwidth]{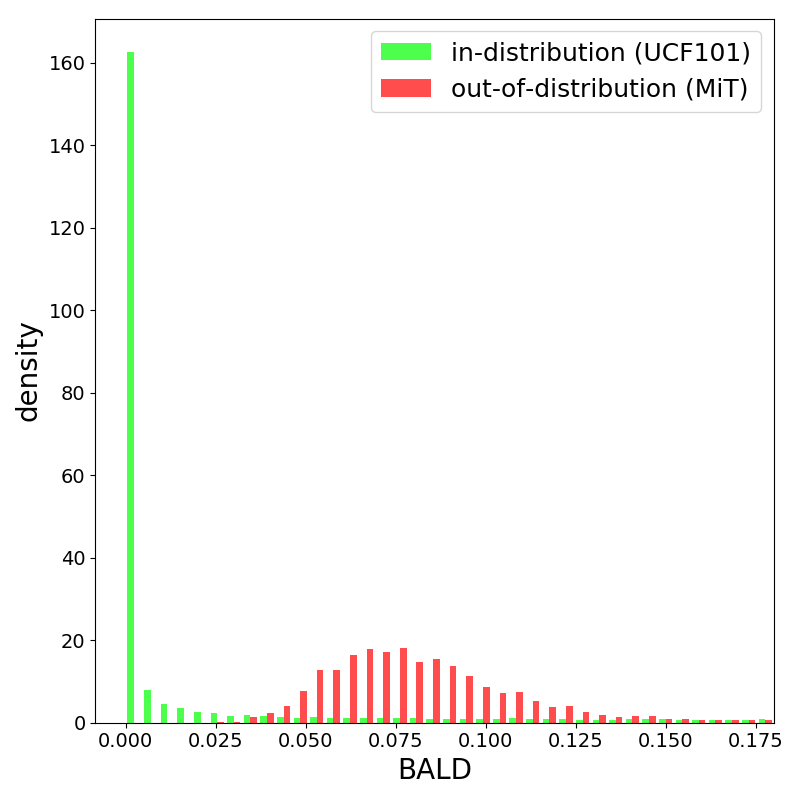}
\caption{\small{BALD (Stochastic VI)}}
\end{subfigure}
\caption{\small{Density histogram of uncertainty measures from Bayesian DNN  MC dropout and stochastic variational inference (Stochastic VI) models for UCF101 dataset as in-distribution samples and MiT datasets as out-of-distribution samples. [The density histogram is a histogram with area normalized to one.]}}
\label{fig:uncertnityestUCF101}
\end{figure}

\section{Conclusions}

We presented a novel Bayesian DNN model for visual activity recognition which uses stochastic variational inference technique during training to infer the approximate posterior distribution around model parameters and perform Monte Carlo sampling on the posteriors. We evaluated our model using a subset of classes from the Moments-in-Time (MiT) dataset. The presented results demonstrate Bayesian DNNs can provide more reliable confidence measures as compared to the conventional DNNs. The uncertainty estimates obtained for wrong predictions and out-of-distribution samples indicate Bayesian DNNs can be more reliable for the visual activity recognition. We envision to extend this framework to multimodal inputs (e.g., audio and vision) where the confidence estimates obtained from Bayesian DNNs can help in fusing information based on uncertainty estimates to achieve better predictions.

\bibliography{BAR_v02}{}

\begin{thebibliography}{28}
\providecommand{\natexlab}[1]{#1}
\providecommand{\url}[1]{\texttt{#1}}
\expandafter\ifx\csname urlstyle\endcsname\relax
  \providecommand{\doi}[1]{doi: #1}\else
  \providecommand{\doi}{doi: \begingroup \urlstyle{rm}\Url}\fi

\bibitem[Bulling et~al.(2014)Bulling, Blanke, and Schiele]{bulling2014tutorial}
Andreas Bulling, Ulf Blanke, and Bernt Schiele.
\newblock A tutorial on human activity recognition using body-worn inertial
  sensors.
\newblock \emph{ACM Computing Surveys (CSUR)}, 46\penalty0 (3):\penalty0 33,
  2014.

\bibitem[Hershey et~al.(2017)Hershey, Chaudhuri, Ellis, Gemmeke, Jansen, Moore,
  Plakal, Platt, Saurous, Seybold, et~al.]{hershey2017cnn}
Shawn Hershey, Sourish Chaudhuri, Daniel~PW Ellis, Jort~F Gemmeke, Aren Jansen,
  R~Channing Moore, Manoj Plakal, Devin Platt, Rif~A Saurous, Bryan Seybold,
  et~al.
\newblock Cnn architectures for large-scale audio classification.
\newblock In \emph{Acoustics, Speech and Signal Processing (ICASSP), 2017 IEEE
  International Conference on}, pages 131--135. IEEE, 2017.

\bibitem[Caba~Heilbron et~al.(2015)Caba~Heilbron, Escorcia, Ghanem, and
  Carlos~Niebles]{caba2015activitynet}
Fabian Caba~Heilbron, Victor Escorcia, Bernard Ghanem, and Juan Carlos~Niebles.
\newblock Activitynet: A large-scale video benchmark for human activity
  understanding.
\newblock In \emph{Proceedings of the IEEE Conference on Computer Vision and
  Pattern Recognition}, pages 961--970, 2015.

\bibitem[Piyathilaka and Kodagoda(2013)]{piyathilaka2013gaussian}
Lasitha Piyathilaka and Sarath Kodagoda.
\newblock Gaussian mixture based hmm for human daily activity recognition using
  3d skeleton features.
\newblock In \emph{Industrial Electronics and Applications (ICIEA), 2013 8th
  IEEE Conference on}, pages 567--572. IEEE, 2013.

\bibitem[Kim et~al.(2010)Kim, Helal, and Cook]{kim2010human}
Eunju Kim, Sumi Helal, and Diane Cook.
\newblock Human activity recognition and pattern discovery.
\newblock \emph{IEEE Pervasive Computing}, 9\penalty0 (1):\penalty0 48, 2010.

\bibitem[Kay et~al.(2017)Kay, Carreira, Simonyan, Zhang, Hillier,
  Vijayanarasimhan, Viola, Green, Back, Natsev, et~al.]{kay2017kinetics}
Will Kay, Joao Carreira, Karen Simonyan, Brian Zhang, Chloe Hillier, Sudheendra
  Vijayanarasimhan, Fabio Viola, Tim Green, Trevor Back, Paul Natsev, et~al.
\newblock The kinetics human action video dataset.
\newblock \emph{arXiv preprint arXiv:1705.06950}, 2017.

\bibitem[Monfort et~al.(2018)Monfort, Zhou, Bargal, Andonian, Yan,
  Ramakrishnan, Brown, Fan, Gutfruend, Vondrick, et~al.]{monfort2018moments}
Mathew Monfort, Bolei Zhou, Sarah~Adel Bargal, Alex Andonian, Tom Yan, Kandan
  Ramakrishnan, Lisa Brown, Quanfu Fan, Dan Gutfruend, Carl Vondrick, et~al.
\newblock Moments in time dataset: one million videos for event understanding.
\newblock \emph{arXiv preprint arXiv:1801.03150}, 2018.

\bibitem[Abu-El-Haija et~al.(2016)Abu-El-Haija, Kothari, Lee, Natsev, Toderici,
  Varadarajan, and Vijayanarasimhan]{abu2016youtube}
Sami Abu-El-Haija, Nisarg Kothari, Joonseok Lee, Paul Natsev, George Toderici,
  Balakrishnan Varadarajan, and Sudheendra Vijayanarasimhan.
\newblock Youtube-8m: A large-scale video classification benchmark.
\newblock \emph{arXiv preprint arXiv:1609.08675}, 2016.

\bibitem[Szegedy et~al.(2017)Szegedy, Ioffe, Vanhoucke, and
  Alemi]{szegedy2017inception}
Christian Szegedy, Sergey Ioffe, Vincent Vanhoucke, and Alexander~A Alemi.
\newblock Inception-v4, inception-resnet and the impact of residual connections
  on learning.
\newblock In \emph{AAAI}, volume~4, page~12, 2017.

\bibitem[Wang et~al.(2016)Wang, Xiong, Wang, Qiao, Lin, Tang, and
  Van~Gool]{wang2016temporal}
Limin Wang, Yuanjun Xiong, Zhe Wang, Yu~Qiao, Dahua Lin, Xiaoou Tang, and Luc
  Van~Gool.
\newblock Temporal segment networks: Towards good practices for deep action
  recognition.
\newblock In \emph{European Conference on Computer Vision}, pages 20--36.
  Springer, 2016.

\bibitem[Zhou et~al.(2017)Zhou, Andonian, and Torralba]{zhou2017temporal}
Bolei Zhou, Alex Andonian, and Antonio Torralba.
\newblock Temporal relational reasoning in videos.
\newblock \emph{arXiv preprint arXiv:1711.08496}, 2017.

\bibitem[Tran et~al.(2015)Tran, Bourdev, Fergus, Torresani, and
  Paluri]{tran2015learning}
Du~Tran, Lubomir Bourdev, Rob Fergus, Lorenzo Torresani, and Manohar Paluri.
\newblock Learning spatiotemporal features with 3d convolutional networks.
\newblock In \emph{Proceedings of the IEEE international conference on computer
  vision}, pages 4489--4497, 2015.

\bibitem[Hara et~al.(2018)Hara, Kataoka, and Satoh]{hara3dcnns}
Kensho Hara, Hirokatsu Kataoka, and Yutaka Satoh.
\newblock Can spatiotemporal 3d cnns retrace the history of 2d cnns and
  imagenet?
\newblock In \emph{Proceedings of the IEEE Conference on Computer Vision and
  Pattern Recognition (CVPR)}, pages 6546--6555, 2018.

\bibitem[Neal(2012)]{neal2012bayesian}
Radford~M Neal.
\newblock \emph{Bayesian learning for neural networks}, volume 118.
\newblock Springer Science \& Business Media, 2012.

\bibitem[Gal(2016)]{gal2016uncertainty}
Yarin Gal.
\newblock Uncertainty in deep learning.
\newblock \emph{University of Cambridge}, 2016.

\bibitem[Kendall and Cipolla(2016)]{kendall2016modelling}
Alex Kendall and Roberto Cipolla.
\newblock Modelling uncertainty in deep learning for camera relocalization.
\newblock In \emph{2016 IEEE international conference on Robotics and
  Automation (ICRA)}, pages 4762--4769. IEEE, 2016.

\bibitem[Kendall et~al.(2015)Kendall, Badrinarayanan, and
  Cipolla]{kendall2015bayesian}
Alex Kendall, Vijay Badrinarayanan, and Roberto Cipolla.
\newblock Bayesian segnet: Model uncertainty in deep convolutional
  encoder-decoder architectures for scene understanding.
\newblock \emph{arXiv preprint arXiv:1511.02680}, 2015.

\bibitem[Kendall et~al.(2017)Kendall, Gal, and Cipolla]{kendall2017multi}
Alex Kendall, Yarin Gal, and Roberto Cipolla.
\newblock Multi-task learning using uncertainty to weigh losses for scene
  geometry and semantics.
\newblock \emph{arXiv preprint arXiv:1705.07115}, 3, 2017.

\bibitem[Welling and Teh(2011)]{welling2011bayesian}
Max Welling and Yee~W Teh.
\newblock Bayesian learning via stochastic gradient langevin dynamics.
\newblock In \emph{Proceedings of the 28th International Conference on Machine
  Learning (ICML-11)}, pages 681--688, 2011.

\bibitem[Graves(2011)]{graves2011practical}
Alex Graves.
\newblock Practical variational inference for neural networks.
\newblock In \emph{Advances in neural information processing systems}, pages
  2348--2356, 2011.

\bibitem[Ranganath et~al.(2013)Ranganath, Gerrish, and
  Blei]{ranganath2013black}
Rajesh Ranganath, Sean Gerrish, and David~M Blei.
\newblock Black box variational inference.
\newblock \emph{arXiv preprint arXiv:1401.0118}, 2013.

\bibitem[Blundell et~al.(2015)Blundell, Cornebise, Kavukcuoglu, and
  Wierstra]{blundell2015weight}
Charles Blundell, Julien Cornebise, Koray Kavukcuoglu, and Daan Wierstra.
\newblock Weight uncertainty in neural networks.
\newblock \emph{arXiv preprint arXiv:1505.05424}, 2015.

\bibitem[Gal and Ghahramani(2016)]{gal2016dropout}
Yarin Gal and Zoubin Ghahramani.
\newblock Dropout as a bayesian approximation: Representing model uncertainty
  in deep learning.
\newblock In \emph{international conference on machine learning}, pages
  1050--1059, 2016.

\bibitem[Blei et~al.(2017)Blei, Kucukelbir, and McAuliffe]{blei2017variational}
David~M Blei, Alp Kucukelbir, and Jon~D McAuliffe.
\newblock Variational inference: A review for statisticians.
\newblock \emph{Journal of the American Statistical Association}, 112\penalty0
  (518):\penalty0 859--877, 2017.

\bibitem[Bishop(2006)]{bishop2006pattern}
Christopher~M Bishop.
\newblock Pattern recognition and machine learning (information science and
  statistics) springer-verlag new york.
\newblock \emph{Inc. Secaucus, NJ, USA}, 2006.

\bibitem[Houlsby et~al.(2011)Houlsby, Husz{\'a}r, Ghahramani, and
  Lengyel]{houlsby2011bayesian}
Neil Houlsby, Ferenc Husz{\'a}r, Zoubin Ghahramani, and M{\'a}t{\'e} Lengyel.
\newblock Bayesian active learning for classification and preference learning.
\newblock \emph{arXiv preprint arXiv:1112.5745}, 2011.

\bibitem[Wen et~al.(2018)Wen, Vicol, Ba, Tran, and Grosse]{wen2018flipout}
Yeming Wen, Paul Vicol, Jimmy Ba, Dustin Tran, and Roger Grosse.
\newblock Flipout: Efficient pseudo-independent weight perturbations on
  mini-batches.
\newblock \emph{arXiv preprint arXiv:1803.04386}, 2018.

\bibitem[Soomro et~al.(2012)Soomro, Zamir, and Shah]{soomro2012ucf101}
Khurram Soomro, Amir~Roshan Zamir, and Mubarak Shah.
\newblock Ucf101: A dataset of 101 human actions classes from videos in the
  wild.
\newblock \emph{arXiv preprint arXiv:1212.0402}, 2012.

\end{thebibliography}
\end{document}